\documentclass[preprint,12pt]{elsarticle}

\usepackage{graphicx}
\usepackage{amssymb}

\usepackage{booktabs} 
\usepackage[utf8]{inputenc}
\usepackage{color}
\definecolor{verde}{rgb}{0.32,0.68,0.31}
\definecolor{orange}{RGB}{255,127,0}
\definecolor{brown}{RGB}{150,70,0}
\definecolor{red}{RGB}{255,90,90}
\definecolor{darkred}{RGB}{150,0,0}
\definecolor{myred}{RGB}{200,50,50}
\definecolor{green}{RGB}{127,255,127}
\definecolor{darkgreen}{RGB}{0,127,0}
\definecolor{mygreen}{RGB}{60,180,60}
\definecolor{lightblue}{RGB}{150,150,255}
\definecolor{blue}{RGB}{127,127,255}
\definecolor{darkblue}{RGB}{0,0,127}
\definecolor{myblue}{RGB}{80,80,200}
\definecolor{grey}{RGB}{127,127,127}
\definecolor{pink}{RGB}{255,180,180}
\definecolor{lightgrey}{RGB}{180,180,180}

\usepackage{graphicx}
\usepackage{subfig}
\usepackage{listings}
\usepackage{multirow}

\usepackage{url}

\usepackage{breakurl}
\usepackage[breaklinks]{hyperref}

\usepackage{listings}
\usepackage{booktabs}
\usepackage{enumitem}

\usepackage{framed}

\newcommand{\sidenote}[1]{}

\newcommand{\sidenoteLidia}[1]{}

\newcommand{\sidenoteSusumu}[1]{}

\usepackage[ruled]{algorithm2e} 

\SetAlFnt{\small}
\SetAlCapFnt{\small}
\SetAlCapNameFnt{\small}
\SetAlCapHSkip{0pt}
\IncMargin{-\parindent}

\makeatletter
\def\ps@pprintTitle{%
   \let\@oddhead\@empty
   \let\@evenhead\@empty
   \let\@oddfoot\@empty
   \let\@evenfoot\@oddfoot
}
\makeatother

\begin{document}

\begin{frontmatter}


\title{General-purpose Declarative Inductive Programming with Domain-Specific Background Knowledge for \\Data Wrangling Automation
}



\author{Lidia Contreras-Ochando, César Ferri, José Hernández-Orallo,\\
Fernando Martínez-Plumed, María José Ramírez-Quintana}
\address{Universitat Politècnica de València, Spain \{liconoc,cferri,jorallo,fmartinez,mramirez\}@dsic.upv.es}

\author{Susumu Katayama}
\address{University of Miyazaki, Japan\\ skata@cs.miyazaki-u.ac.jp}

\begin{abstract}
{\footnotesize
Given one or two examples, humans are good at understanding how to solve a problem independently of its domain, because they are able to detect what the problem is and to choose the appropriate background knowledge according to the context. For instance, presented with the string ``8/17/2017" to be transformed to ``17th of August of 2017", humans will process this in two steps: (1) they recognise that it is a date and (2) they map the date to the 17th of August of 2017. Inductive Programming (IP) aims at learning declarative (functional or logic) programs from examples. Two key advantages of IP are the use of background knowledge and the ability to synthesise programs from a few input/output examples (as humans do). In this paper we propose to use IP as a means for automating repetitive data manipulation tasks, frequently presented during the process of {\em data wrangling} in many data manipulation problems. Here we show that with the use of general-purpose declarative (programming) languages jointly with generic IP systems
and the definition of domain-specific knowledge, many specific data wrangling problems from different application domains can be automatically solved from very few examples. We also propose an integrated benchmark for data wrangling, which we share publicly for the community.
}
\end{abstract}


\end{frontmatter}

\section{Introduction}

 
The term `data wrangling' usually refers to a great deal of repetitive and very time-consuming data preparation tasks, such as the acquisition, integration, manipulation, cleansing, enriching and transformation of data from their raw format to a more structured and valuable form for easy access and analysis \cite{kandel2011research}. The use of ETL\footnote{Originally from the data warehousing terminology, ETL is the process responsible for the extraction, transformation and load of the data into a repository.} 
tools and other scripting languages for data wrangling partially alleviate the problem, but most of the effort is still manual and non-systematic. Consequently, progress in the (semi-)automation of data wrangling tasks can have an enormous impact in the costs of data science projects and other data manipulation problems, and can also allow data scientists focus on the valuable knowledge discovery process or in the actual task they are doing.

Many data wrangling problems look automatable, especially because the user can indicate a few illustrative examples that can be used by an Inductive Programming (IP) system \cite{flener1999inductive,muggleton1994inductive,kitzelmann2006inductive,hernandez1999strong} to infer a pattern, or inductive hypothesis, that can be used to complete the rest of the examples automatically. Table~\ref{tab:wrangling-example1} shows one example  that can be completed by non-expert people easily, without further knowledge about the source of the data. It is a very encapsulated problem, inputs and outputs, which should be well handled by machines. 

\begin{table}[!h]
	\centering
	{
		\begin{tabular}{c|c|c|c|c|c|c}
			
			\textbf{Id} & \textbf{Input} & \textbf{Outputs} & $\:\:\:\:\:$ & \textbf{Id} &\textbf{Input} & \textbf{Output}\\ \hline
			\hline
			1 & 25-03-74 & \textit{25/03/74} & & 5 & 17-05-17 & 17/05/17 \\ 
			2 & 29-03-86 & \textit{29/03/86} & & 6 & 25-08-05 &  25/08/05 \\
			3 & 11-02-96 &  11/02/96 & & 7 & 30-06-75 & 30/06/75\\
			4 & 11-17-98 &   17/11/98 & & 8  & ...      &  ... \\ 
	
			\hline 
            \end{tabular}%
	}

	\caption{Dataset composed of dates (input) and desired output format. An automatic data wrangling system is fed with the two first examples (in italics) and should automatically complete the rest of the cells (outputs).}
	\label{tab:wrangling-example1}
\end{table}

Nevertheless, many other data wrangling problems are more challenging, and require an important degree of background knowledge because they depend on the application context of the data. Table \ref{tab:wrangling-example2} 
shows an example of a common data wrangling problem: given a list of dates, extract the day from each of them. The difficulty in this problem lies in the different date formats (dependent on the original country where the data come from), where the day can be the first, second or third number, and these numbers can be delimited by different symbols. A system based on string transformations may never find the right solution using its search space since it does not know what the real problem is: extracting the first number? the first two digits? or everything before any symbol? In order to understand and complete the transformation, we must know how dates work, their constraints and how they are usually represented. We know that there are only twelve months, that days can only range between 1 and 31 and that years are usually abbreviated with two single digits. 

\begin{table}[!h]
	\centering
	{
            \begin{tabular}{c|c|c|c|c|c|c}
			
			\textbf{Id} & \textbf{Input} & \textbf{Output} & $\:\:\:\:\:$ & \textbf{Id} &\textbf{Input} & \textbf{Output}\\ \hline
			\hline
			1 & 25-03-74 & 25 & & 5 & 17/05/57 & 17 \\ 
			2 & 03/29/86 & 29 & & 6 & 25-08-05 & 25\\
			3 & 21.02.98 & 21 & & 7 & 06 30 1975 &  30 \\
			4 & 1998/12/25 &  25 & & 8  & ...      &  ... \\ 
	
			\hline 
            \end{tabular}%
	}

	\caption{Dataset composed of dates under very different formats (input) from which the day is extracted (output).}
	\label{tab:wrangling-example2}
\end{table}

In order to solve this problem, we can split the data wrangling problem into two steps: first, we need to know which the domain is (e.g., dates); and, second, we need to know which transformations we have to apply to the input to obtain the output. For humans this is a relatively easy step because we have information of the context, but it is not so easy for machines. We need to specify relevant background knowledge as well as the necessary transformations (depending on the domain). Of course, some of this knowledge may be insufficient to sort out some ambiguities, such as ``11.02.18'' (this date can be in DDMMYY, YYMMDD or MMDDYY formats). This problem may be automatically solved by computers (through program synthesis) if they are able to recognise the domain (i.e., dates), and have a sufficiently rich set of functions to deal with the context. Not only does this impose a strong bias that guides the process of finding the transformation pattern that has to be applied but also introduces some useful functions that render the solution (the  inferred program) much shorter in terms of the functions involved. This size of the solution (in terms of primitives/functions involved) is known as the {\em depth} ($d$).

			
	


Of course, dates are not the only kind of data. If we want to deal with physical addresses, we need to provide functions that handle symbols such as ``St'', ``Rd'', order of postcodes, etc. Similarly, if we want to deal with people names, we should understand strings such as ``Mrs'', ``Dr'', etc. Since all of these cases are very common in databases and other kinds of data that are processed in data science projects, we can add the relevant functions to a general domain library. However, as more kinds of data are required, this library would become huge. Even if the depth would have not changed for the original date problem, the inductive inference process needs to choose from a much larger space of functions, which makes it much harder. This is known as the {\em breadth} ($b$). Clearly, both the depth and the breadth highly influence the hardness of the problem, jointly with the number of examples, $n$. Actually, for hypothesis-oriented induction, hardness strongly depends on $d$ and $b$, in a way that is usually exponential, $O(b^d)$ \cite{henderson2009incremental,hernandez2013deep}. How can we keep both, and especially $b$, at very low levels? 

In this paper, we control the depth and breadth of the inductive inference problem by choosing a \textit{domain-specific background knowledge} (DSBK) for each kind of problem. Based on any IP system, which is hypothesis-oriented rather than data-oriented, we see that the effort only depends on these two parameters, $d$ and $b$, being almost constant on the number of examples. The user just needs to suggest which domain to use for a particular problem: dates, times, emails, names, phones, etc. Nevertheless, we envisage that this step is easily automatable too, using some domain inference process that can suggest this to the user, as we discuss at the end of the paper. 
It is important to remark that the inductive inference engine is the same, independently of whether we are handling dates or telephone numbers. We do not build a data wrangling system specialised for a particular domain-specific language for each case. Instead of this, we allow the system to use different DSBKs.

There are several advantages of this approach. The same data wrangling tool can be used for a diversity of problems and domains, without specialised tools for every domain. Second, a set of DSBK libraries can be provided by the tool but also extended by users and communities, especially if the language for adding or modifying functions is general-purpose and well known (e.g., Haskell \cite{jones2003haskell,jones1996compiling}, Prolog \cite{lloyd1987foundations}, etc.). 

Overall, this paper contains several contributions:

\begin{itemize}\itemsep-0.1em 
	\item We illustrate how the use of general program induction, as a kind of hypothesis-driven machine learning, can be applied flexibly for problems where knowledge is important, through the definition of domain-specific domain. More generally, we see a combination of knowledge-based and learning AI approaches.
    \item We show that this generic approach, combining an off-the-shelf IP system with appropriate operators to define the necessary background knowledge for a domain, is able to improve the results of other state-of-the-art --and more specific-- data wrangling approaches.
	\item We analyse how the breadth, depth and number of instances affect the efficiency, showing that we can achieve a trade-off between breadth and depth, and still solve many problems using only one example.
	\item We provide a set of datasets specifically designed to be the first benchmark for the evaluation of data wrangling tools focusing on column transformation problems. 
\end{itemize}

The paper is organised as follows. Section~\ref{sec:related} summarises the most relevant related works. Section~\ref{sec:approach} addresses the problem of automating data wrangling with an IP system. The domains employed are detailed in Section~\ref{sec:domains}. The experimental evaluation is included in Section~\ref{sec:experiments}. Finally, Section~\ref{sec:conclusion} remarks the conclusions and future work.

\section{Related Work}\label{sec:related}


The importance of data wrangling in the quality and cost of data science projects has motivated an enormous effort in techniques and approaches, including commercial platforms that go beyond ETL tools. For instance, \emph{OpenRefine}\footnote{OpenRefine: \url{http://www.openrefine.org}} provides a set of built-in operators to specify data transformations.  
\emph{Ajax} \cite{ajax} brings a SQL-like language to statements extended with advanced facilities for entity resolution) that enables the user to specify the sequence of data transformations. Trifacta \emph{Wrangler} \cite{wrangler} generates a ranked list of suggested transformations and text extractions also inferred automatically from user input, the data type and some heuristics using programming-by-demonstration techniques. Dataxformer \cite{abedjan2015dataxformer} uses a big corpus of web tables of data to find the most useful transformation for each problem.

The above systems are able to use different approaches depending on the data type. 
In general, these tools have predefined ``types'' or structures for emails, gender, phones, credit cards, social security number, etc. However, their pattern generation engines are usually based on predefined rules, but not on a proper AI technology, such as a fully inductive inference system.

Given this limitation
, research has focused on the inductive inference part of the problem, when the pattern involves the combination of several manipulation functions. A new generation of approaches is based on \emph{Inductive Programming}, which has recently shown a large potential for  data wrangling automation \cite{gulwani2014inductive}. IP addresses the problem of learning small (but complex) programs from very few representative input/output examples, generated as the user transforms one (or very few) particular instances of the data. The application has been so successful that Microsoft includes some of these tools in Excel, known as \emph{FlashFill} \cite{Gulwani:2011}. 
 One of the reasons of the success of these systems is the use of domain-specific languages (DSLs). As an example, \emph{FlashFill} is able to make syntactic transformations of strings using restricted forms of regular expressions, conditionals and loops on spreadsheet tables.

The use of DSLs has overcome the limitations of general-purpose IP systems such as Progol \cite{mug95}, IGOR2 \cite{kitzelmann2008igor2},  \emph{MagicHaskeller} \cite{katayama2012analytical}, FLIP \cite{Ferri-RamirezHR01}, 
\emph{Metagol} \cite{muggleton2015meta}, \emph{gErl} \cite{Fmartinezphdthesis},  
and many others. The languages behind these systems (Prolog, Haskell, etc.) have such a diversity of functions and possible combinations that the breadth and depth for the search problem is usually problematic. DSLs, on the other hand, are much more ad-hoc when dealing with specific data wrangling problems, and reduce the search space considerably. 

However, the use of DSL systems for data wrangling automation also brings some disadvantages: (1) DSLs imply the use of languages that are specifically defined for a particular type of data processing (e.g., string processing, number processing, etc.). Whenever a new application or domain is required, a new domain-specific language has to be created, and the inductive engine recoded for it. Despite the effort of making this process more efficient in the recent years, it still depends on languages that are not of general purpose, and users need some effort to understand them
; (2) these systems work using a specific set of transformations depending on the type of input, assuming the input to be in a unique format. This means that, even when the domain of the data is known or can be inferred, whether different formats of a data type appear in different rows of the same column (such as the examples of dates seen in Table \ref{tab:wrangling-example2}), the system is unable to find a solution for transforming each example, resulting in the correct transformation of only those examples with the same type of the example used as input.

In this paper, we propose the use of IP systems which (1) are `specialised' with appropriate libraries that define domain-specific background knowledges (DSBKs), reducing the breadth of the search problem; and (2) are able to extract or transform data from one or few input examples to correct outputs, depending on the data domain and context and independently of the different formats appearing on the input column.

\section{Automating data wrangling}\label{sec:approach}

\begin{figure*}[t]
	\centering
	\includegraphics[width=\textwidth]{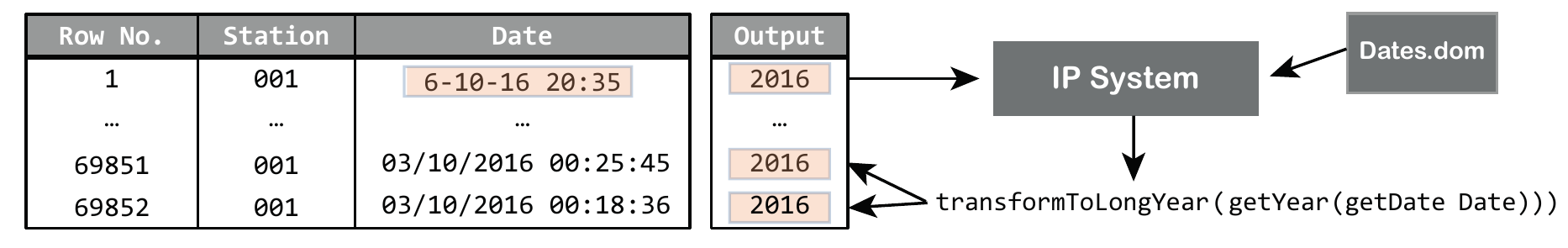}
	\caption{Automating data wrangling with IP: process example. The first row (Data and Output) is used as a input predicate for the IP system. The function returned using the correct domain is applied to the rest of the instances to obtain the outputs.}
	\label{fig:mw-idea}
\end{figure*}

The overall idea is to automate the process of transforming data from one format to another, depending on the data domain, using a general-purpose IP system at the core, but enhanced to handle configurable function libraries for each domain (see Figure \ref{fig:mw-idea}). For this, we do the following steps:

\begin{enumerate}
	\item We take a dataset of input-output pairs and detect the domain of the data.
	\item We set the domain as background knowledge for the IP system.
	\item One or more examples are sent to the IP system as inputs, such as the few first rows in Table \ref{tab:wrangling-example1}, in the same way a user could complete a few examples. These examples are used as input predicates for the system.
	\item With the correct DSBK, The system is able to return a list of transformations addressing the problem as the resulting function ($f$) that is applied to the rest of the inputs, obtaining the new values for the output column. 
\end{enumerate} 

\subsection{Domain-Specific Induction}
For the purpose of this work we have used \emph{MagicHaskeller} \cite{katayama2012analytical} as the IP core system for several reasons\footnote{Note that all the experiments could be replicated using any other IP learning system.}. First of all, \emph{MagicHaskeller} is a general-purpose learning system that works with Haskell, a functional programming language that makes it much easier to add domains and transformations. Besides, \emph{MagicHaskeller} is a very powerful system that can solve many problems using only one example from the data. It is also possible to provide \emph{MagicHaskeller} with different data wrangling domains as different sets of background knowledge's functions. 

In a nutshell, \emph{MagicHaskeller} is a general-purpose inductive functional programming system that learns Haskell programs from pairs of input-output examples, also expressed in Haskell. \emph{MagicHaskeller} receives an input example ($x$) and the expected result ($y$), and returns a list of functions ($f$) that make the values of the expressions $f x$ and $y$ be equal, which in Haskell notation is expressed as the boolean predicate $\texttt{f x == y}$. \emph{MagicHaskeller} looks for combinations of one or more functions that are defined in its library to work like the $f$ above. 




\emph{MagicHaskeller} works in two steps: \textit{(1)} The \emph{Hypotheses Generation} phase, and \textit{(2)} the \emph{Hypotheses Selection} phase. In (1), \emph{MagicHaskeller} starts with a predefined $d_{max}$ value (maximum $d$ allowed for the solution) and a set of $b$ functions in the library. Then, \emph{MagicHaskeller} continues with the preparation of hypotheses by generating all the type-correct expressions that can be expressed by function application and lambda abstraction using up to the maximum depth ($d_{max}$) the functions provided in the library. Although \emph{MagicHaskeller} is very powerful for finding the simplest and most effective solutions (that is, those with smallest Kolmogorov complexity), depending on the problem, the solution might require the combination of many function symbols (that is, a solution with a large depth $d$). When the $d$ required is higher than the $d_{max}$ value used, \emph{MagicHaskeller} is not able to find the solution (because it cannot reach the necessary number of functions combined). Trying to increase the $d_{max}$ value to achieve the result may cause an increment of time over the top. On the contrary, trying to reduce $d$, we may be tempted to add many powerful and abstract functions to the library. But, in this case, \emph{MagicHaskeller} will have too many primitives to choose from (the breadth value $b$), and may not find it either because of the time needed to combine all of them. 

\begin{figure}[h!]%
    \centering
    \subfloat[]{{\includegraphics[width= 0.75\textwidth]{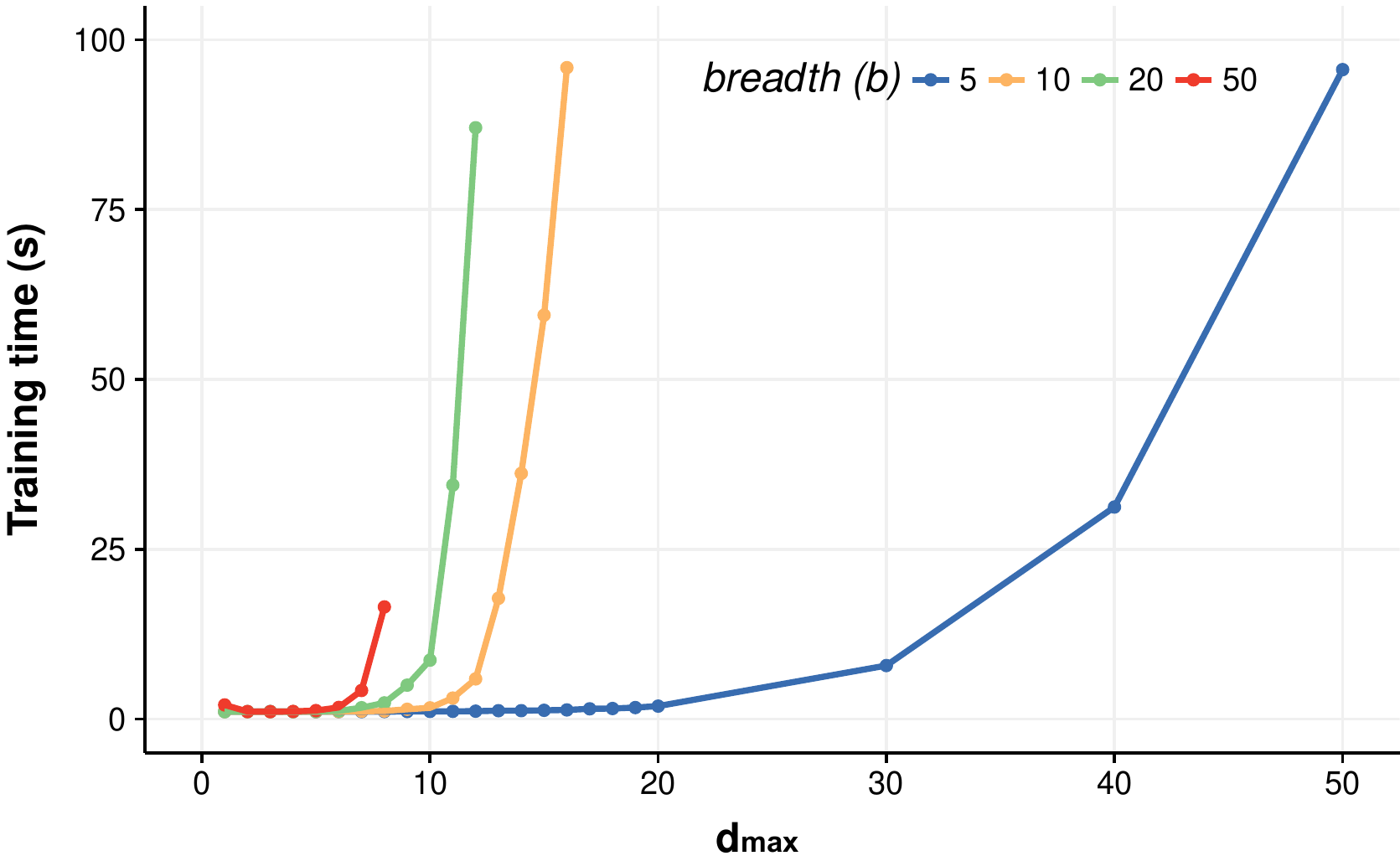} }}%
    \qquad
    \subfloat[]{{\includegraphics[width= 0.75\textwidth]{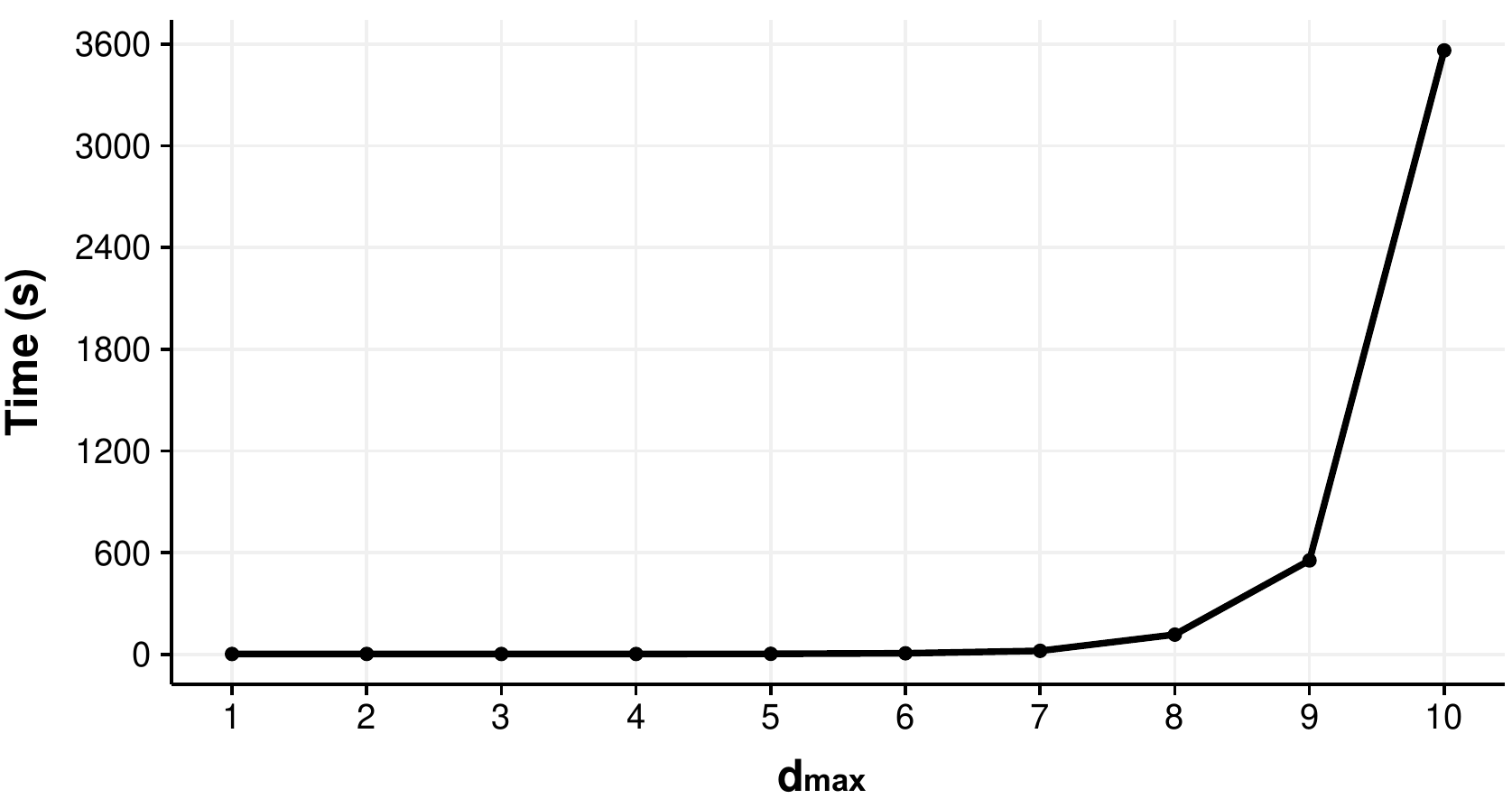} }}%
    \caption{\textbf{(a)} Time \emph{MagicHaskeller} needs for training with a set of primitives depending on the maximum number of primitives that are allowed in any synthesised function ($d_{max}$) and the number of primitives in the set ($b$). \textbf{(b)} Time \emph{MagicHaskeller} needs for training and solve the same problem (concatenate two strings), using a set fixed of $b=15$ primitives, with varying $d_{max}$ from 1 to 10.}%
    \label{fig:mh-performance}%
    \vspace{0.4cm}
\end{figure}

In general, it is usually estimated that for hypothesis-driven inductive inference, the computational complexity might be in the order of $O(b^d) $ \cite{gottlob1997complexity}. Figure \ref{fig:mh-performance} (a) illustrates this by showing the time used by \emph{MagicHaskeller} in this phase when we vary both the number of functions included in the library ($b$) and the maximum depth value to obtain the solution ($d_{max}$).

Finally, in phase (2) we can provide one or more examples (as I/O pairs) to solve a specific problem. \emph{MagicHaskeller} will use the combinations learnt at (1) to find one or more possible solutions to the problem. This solution (if exists) will be a combination of $d$ functions (where $d \leq d_{max}$). In this regard, Figure \ref{fig:mh-performance}(b) shows the time spent during phases (1) and (2) to solve an specific problem (with actual solution of $d=1$), using the same set of functions (with $b=15$), but changing the $d_{max}$ value. 
We acknowledge that $d_{max}$ value has a strong influence too even when there are solutions that require fewer primitives than the maximum depth. Given the heuristics and optimisations included in \emph{MagicHaskeller}, it is still possible to have solutions in cases where $O(b^d)$ grows very fast, but we still see the exponential behaviour in both cases. In the next sections we will show that a good trade-off between $d$ and $b$  can be achieved by using specific domain libraries. Thus, in that follows, we will refer to our approach as \emph{Domain-Specific Induction (DSI)}.


\subsection{Domain-specific Background Knowledge}\label{sec:domains}


By default, \emph{MagicHaskeller} includes a list of 189 basic Haskell functions. Table \ref{tab:mh-functions} shows some of these functions\footnote{The complete list of functions is published at: \url{https://github.com/liconoc/DataWrangling-DSI}}. Although \emph{MagicHaskeller} is able to solve many string and boolean problems by using its default library \cite{katayama2012analytical}, this list of functions is not enough to solve more complex problems. For instance, the example shown previously in Table \ref{tab:wrangling-example1} is impossible to solve with \emph{MagicHaskeller}'s default library since there is a need to replace each dash symbol ('-') with a slash symbol ('/'), and \emph{MagicHaskeller} is unable to generate or use any character or digit if it is not defined as constant in its library or if it is not provided as an input parameter. 

\begin{table*}[h!]
\centering
\resizebox{\textwidth}{!}{%
		\begin{tabular}{l|l}
			
			\multicolumn{2}{ c }{Functions} \\ \hline
			\hline		          
            0 :: Int  &
            1 :: Int \\
            (++) :: forall a . (->) ([a]) ([a] -> [a])   &
            filter :: forall a . (a -> Bool) -> [a] -> [a] \\
            isLower :: (->) Char Bool  &
            words :: [Char] -> [[Char]] \\
            (+) :: Int -> Int  &
            True :: Bool \\
            False :: Bool  &
            isPunctuation :: (->) Char Bool \\
            (+) :: (->) Int ((->) Int Int)  &
            takeWhile :: forall a . (a -> Bool) -> [a] -> [a] \\
            isDigit :: (->) Char Bool  &
            not :: (->) Bool Bool \\
            (-) :: Int -> Int -> Int  &
            (\&\&) :: (->) Bool ((->) Bool Bool) \\
            (||) :: (->) Bool ((->) Bool Bool)  &
            not :: (->) Bool Bool \\
            (-) :: Int -> Int -> Int  &
            reverse :: forall a . [a] -> [a]\\
            \hline 
            
            \end{tabular}%
        }  
\caption{Some default functions in \emph{MagicHaskeller}.}
	\label{tab:mh-functions}
\end{table*}

In order to solve this kind of problem we have to add constants to the library and some new functions to work with string problems. For this particular case, we can solve the problem by adding the primitives in Table \ref{tab:punctuation-functions} to the library.

\begin{table*}[h!]
\centering
\resizebox{\textwidth}{!}{%

		\begin{tabular}{l|l}
			
			Functions & Description \\ \hline
			\hline		          
            dash :: [Char]  & Constant for dash ('-') symbol\\
            slash :: [Char]   & Constant for slash ('/') symbol  \\
            changePunctuationString :: [Char] -> [Char] -> [Char]  & Replace a punctuation sign\\
			\hline 
            
            \end{tabular}%
        } 
\caption{Functions needed to replace a dash symbol with a slash symbol using \emph{MagicHaskeller}.}
	\label{tab:punctuation-functions}
\end{table*}

Following this and some other examples \cite{nishida_7_2016} and the most common operators used by other data science tools \cite{noauthor_key_nodate}\cite{wrangler}\cite{noauthor_wrangle_nodate}, we have added to \emph{MagicHaskeller} many new functions for solving common problems related to string manipulation. Concretely, we have added 108 functions to solve the following string operations:

\begin{itemize}
	\item \textbf{Constants:} Symbols, numbers, words or list of words.
    \item \textbf{Map:} Boolean functions for checking string structures.	
	\item \textbf{Transform:} Functions that return the string transformed using one or more of the following operations:
    	\begin{itemize}
        \item \textbf{Add:} Appending elements to a string, adding them at the beginning, ending or a fixed position.
        \item \textbf{Split:} Splitting the string into two or more strings by positions, constants or a given parameter.	
        \item \textbf{Concatenate:} Joining strings, elements of an array, constants or given parameters with or without adding other parameters or constants between them.
        \item \textbf{Replace:} Changing one or more string elements by some other given element 
        . This operation includes converting a string to uppercase and lowercase.
        \item \textbf{Exchange:} Swapping elements inside strings.
        \item \textbf{Delete/Drop/Reduce:} Deleting one or more string elements by some other given parameter, a position, size or mapping some parameter or constant.
        \item \textbf{Extraction:} Get one or more string elements.
    	\end{itemize}
\end{itemize} 

\noindent 
With this set of functions in the system's library, we are able to solve many common string manipulation problems, such as the example in Table \ref{tab:wrangling-example1}. However, the results can be less accurate for different examples. Trying to solve the example in Table \ref{tab:wrangling-example2} using the first row as a predicate (f "25-03-74" == "25") the first three results obtained may be: (1) \emph{takeWhile isDigit "25-03-86"}; (2) \emph{getStartToFirstSymbol "25-03-86" dash}; and (3) \emph{take 2 "25-03-86"}. When we apply these functions to the firs row, we obtain the desired results, but, what happens if we apply these functions to the rest of the table? We can see the results in Table \ref{tab:bad-functions-example}. It should be noted that only in the cases when the day is the first element of the date (with solutions 1 and 3) and the next symbol is a dash (with solution 2) the result is correct. The problem here is that we cannot assume that all the data in a column has always the same format. In this case, dates come from very different formats and extracting the first element not always results in getting the day. When data belong to a particular domain and the problem at hand ends up being a very exclusive task pertaining to that domain, more precise functions are needed in order to get correct results considering the context. 
However, as we have seen  in the previous section,  it is critical to reduce $b$ while at the same time having the appropriate abstract primitives to learn the function with a short hypothesis (small $d$). This could be solved by  detecting the domain of the data to be transformed and choosing a domain-specific library for it.

\begin{table*}[h!]
	\centering
	{
	\resizebox{\textwidth}{!}{%
		\begin{tabular}{c|c|c|c|c|c}
			
			\textbf{Id} & \textbf{Input} & \textbf{Expected Output} & \textbf{Actual Output (1)} & \textbf{Actual Output (2)} & \textbf{Actual Output (3)} \\ \hline
			\hline
			1 & \emph{25-03-74} & \emph{25} &  &  &  \\
            2 & 03/29/86 & 29 & \textcolor{red}{03} & \textcolor{red}{03/29/86} & \textcolor{red}{03}  \\
            3 & 21.02.98 & 11 & \textcolor{darkgreen}{21} & \textcolor{red}{21.02.98} & \textcolor{darkgreen}{21} \\
            4 & 1998/12/25 &  25 & \textcolor{red}{1998} & \textcolor{red}{1998/12/25} & \textcolor{red}{19} \\
            5 & 17/05/57 & 17 & \textcolor{darkgreen}{17} & \textcolor{red}{17/05/17} & \textcolor{darkgreen}{17}\\ 			
            6 & 25-08-05 & 25 & \textcolor{darkgreen}{25} & \textcolor{darkgreen}{25} & \textcolor{darkgreen}{25}\\			
            7 & 06 30 1975 &  30 & \textcolor{red}{06}  & \textcolor{red}{06 30 1975} & \textcolor{red}{06}\\			
            8  & ...  &  ... & ... &  ... & ... \\ 	
			\hline 
            \end{tabular}%
        }
	}

	\caption{Example of a dataset with an input column composed of dates under very different formats, the expected output (day) and the actual outputs obtained using an inductive system with string manipulation functions. The first row is used as input predicate for the system. Green examples are correct results. Red examples are incorrect results. Solution (1): \emph{takeWhile isDigit "input"}; Solution (2): \emph{getStartToFirstSymbol "input" dash}; and Solution (3): \emph{take 2 "input"}.}
	\label{tab:bad-functions-example}
\end{table*}

A high number of different domains can appear in any data science project related to data manipulation problems. In order to make our experiments and as other data wrangling tools have already done \cite{PottersWheel}\cite{noauthor_supported_nodate}\cite{noauthor_data_nodate}, 
we have selected some of the most used domains \cite{noauthor_supported_nodate} and their most common problems \cite{PottersWheel} to work with. In this sense, for each domain we have a different background knowledge with a set of possible transformations. As we are working with \emph{MagicHaskeller}, they are represented as Haskell functions.
These are independent text files, editable by the user, which can be included as a parameter when \emph{MagicHaskeller} is invoked. The DSBK files are:

\begin{itemize}
 \item \textbf{Dates}: The DSBK includes 23 functions related to date manipulation (and includes 139 primitives from the freetext BK), such as determining whether a substring is a month, getting the day in ordinal format, converting a month to numeric format or extending a two-digit year to a four-digit full format. 
 \item \textbf{Emails}: This DSBK includes 9 functions related to email manipulation (and includes 93 primitives from the freetext BK), such as getting the words after or before the '@' symbol, append the '@' symbol at the end of a string or join two strings with the '@' symbol. 
 \item \textbf{Names}: The DSBK includes 12 functions related to personal names manipulation (and includes 104 primitives from the freetext BK), such as getting the initials of a name or creating a user login. 
    \item \textbf{Phones}: This DSBK includes 5 functions related phone numbers manipulation (and includes 124 primitives from the freetext BK), for example, setting the prefix by a country name or code. 
 \item \textbf{Times}: This DSBK has 24 functions to deal with strings containing time (and includes 124 primitives from the freetext BK), such as 12/24h formats or changing time zone.
\end{itemize}

\noindent 
In total we have used 374 different functions. Although we are considering only six domains besides the basic string manipulation functions, it should be noted that
many other domains can be created, and it is also easy to build domains that are defined as the union between two existing domains. Also, \emph{MagicHaskeller} can be called with a small $d_{max}$ parameter with one domain to get results quickly and, if unsuccessful, try with a larger $d_{max}$ or another domain. In this way, the search effort can be better handled, depending on the knowledge of the domain and the expected size of the solution.

\section{Experiments}\label{sec:experiments}

The aim of our experiments is to analyse the extended capabilities of an IP learning system as a data wrangler. Besides, the experiments explore the improvement in the results when selecting the right DSBK in front of using a general background knowledge or an inappropriate DSBK. Also, and more importantly, we want to compare with other data wrangling systems on a range of data wrangling problems. 

To perform the experiments we have followed a trained/test evaluation procedure, similar to \cite{bhupatiraju2017deep,Ellis2017,singh2016transforming,Gulwani:2011,singh2015predicting}. We have used a set of datasets with different data wrangling problems (explained in the following sub-section) including inputs and expected outputs. For each of these datasets, we use only the first example as the input predicate for the IP system.  Then, we fed the system with this first input/output example using, for each dataset, all the different DSBK. The result is a function $f$ that is applied to the rest of the outputs. The accuracy in each case is the result of compare the transformed outputs with the real expected outputs.

For replicability reasons, the source code (scripts, domain files, primitive files \emph{MagicHaskeller}, etc.) and all the results of these experiments are available online \footnote{\url{https://github.com/liconoc/DataWrangling-DSI}}.

\subsection{Data Wrangling Benchmark}

Unfortunately, there is no general benchmark or public dataset repository accessible in reusable formats to analyse the quality of new data wrangling tools (for instance, in \cite{Ellis2017} the authors use a dataset with hundreds of data manipulation problems, but the benchmark is not public). In order to overcome this limitation, we have collected most of the datasets tested previously in other tools for data manipulation (such as \emph{FlashFill} or \emph{Wrangler}) 
and presented in the literature \cite{bhupatiraju2017deep,Ellis2017,singh2016transforming,Gulwani:2011,singh2015predicting}. In addition, we have generated new datasets based on problems from these papers.

\begin{table*}[!ht]
\centering
\resizebox{\textwidth}{!}{%
\begin{tabular}{clcl}
				
				\textbf{id} & \textbf{Domain} &\textbf{\#Ex.} & \textbf{Description of the problem to solve}  \\ 
				\hline \hline
                1 & Freetext & 12 & Complete brackets (From \cite{Ellis2017})  \\ 
                2 & Freetext & 12 & Extract the first character (From \cite{contrerasgeneral})  \\ 
                3 & Freetext & 24 & Delete punctuation (From \cite{Ellis2017}) \\
                4 & Freetext & 18 & Extract the capital letters (From \cite{Ellis2017}) \\
                5 & Freetext & 12 & Extract a substring (From \cite{polozov2016program}
                ) \\ 
                \hline
                6 & Dates & 26 & Change the punctuation of a date (From \cite{singh2016transforming}) \\ 
                7 & Dates & 26 & Extract the day from a date (Generated) \\ 
                8 & Dates & 12 & Extract the day from a date in ordinal format (Generated) \\ 
                9 & Dates & 12  & Extract the month from dates (Generated) \\ 
                10 & Dates & 12  & Extract the name of the month from dates 
                (From \cite{polozov2016program}
                ) \\ 
                11 & Dates & 9 & Add punctuation to a date (From \cite{polozov2016program}
                ) \\ 	
                12 & Dates & 25 & Change date format and punctuation (Generated) \\ 	
                13 & Dates & 12 & Add punctuation and change the format of a date (From \cite{polozov2016program}
                ) \\ 	
				\hline
                14 & Emails & 24 & Extract words after '@' (From \cite{polozov2016program}
                )\\ 
                15 & Emails & 18 & Join words with '@' (From \cite{bhupatiraju2017deep})\\ 
                \hline
                16 & Names & 12 & Generate a login from a name (Generated)\\ 
                17 & Names & 12  & Reduce name from one input (From \cite{Gulwani:2011}) \\ 
				18 & Names & 12 & Reduce name from two inputs (From \cite{Gulwani:2011}) \\ 
                19 & Names & 12 & Extract the honorific forms (From \cite{Gulwani:2011}) \\ 
                \hline
                20 & Phones & 12 & Add phone prefix by country name (From PROSE
                ) \\ 
                21 & Phones & 12 & Add phone prefix by country name and '+' symbol (Generated) \\ 
                22 & Phones & 12 & Add a given phone prefix  (From \cite{polozov2016program}
                ) \\  
                23 & Phones & 12 & Extract a phone number from a string (From \cite{polozov2016program}
                ) \\ 
                24 & Phones & 12 & Add punctuation to a phone number (Generated) \\ 
                \hline
                25 & Times & 12 & Extract the time from a string (Generated) \\ 
                26 & Times & 12 & Append a specific given time (minutes or seconds) (Generated) \\ 
                27 & Times & 12 & Increase the hour by a given value (Generated) \\ 
                28 & Times & 12 & Convert the time to 24h format (Generated) \\ 
                29 & Times & 12 & Convert time by a given time zone (Generated) \\ 	
                \hline
                30 & Units & 12 & Extract the units of a value (From \cite{contrerasgeneral}\\ 
                31 & Units & 12 & Detect the system units by the units of a value (Generated)  \\ 
                32 & Units & 12 & Convert a value to a different unit (Generated)  \\ 
				\hline
			\end{tabular}
            }
\caption{Datasets included in the new data wrangling repository offered for the data science research community. \textit{\#Ex}. shows cardinality.
 \textit{Freetext} represents the functions created for solving string manipulation problems.}\label{tab:datasets}
\end{table*}

Overall, we have collected or generated 32 datasets and we have published them on the first data wrangling dataset repository, which is online and available at \url{http://dmip.webs.upv.es/datawrangling/}. Table~\ref{tab:datasets} shows a summary of the datasets in this new repository.

\subsection{Results}

With a focus on our system, Table~\ref{tab:mainresults} shows the results (accuracy) for all the datasets, using just one example (the first one of each dataset), when \emph{MagicHaskeller} is run without extra DSBK (\textit{default}), adding the string manipulation functions (\textit{freetext}), with a particular DSBK (\textit{dates, emails, names, phones, times, units}) and with all DSBKs together (a unique set of primitives with all the functions together). In each case, \emph{MagicHaskeller} returns a potential solution (or nothing if the problem cannot be solved) which is applied to the rest of input examples to see whether the obtained output matches the expected one. Time execution is limited to $120s$ with $d_{max}=4$. 

\begin{table*}[!ht]
\centering
\resizebox{\textwidth}{!}{%

\begin{tabular}{rcccccccccc}
			
			\textbf{id}& $Domain$ &  $default$ & $freetext$ & $dates$ & $emails$ & $names$ & $phones$ & $times$ & $units$  & $all$ \\ 
		\hline
  1 & freetext & 0.00 & \textbf{1.00} & 0.00 & 0.00 & 0.00 & \textbf{1.00} & \textbf{1.00} & \textbf{1.00} & 0.00 \\ 
  2 & freetext & \textbf{1.00} & \textbf{1.00} & \textbf{1.00} & \textbf{1.00} & \textbf{1.00} & \textbf{1.00} & \textbf{1.00} & \textbf{1.00} & 0.00 \\ 
  3 & freetext & 0.48 & \textbf{1.00} & \textbf{1.00} & \textbf{1.00} & \textbf{1.00} & \textbf{1.00} & \textbf{1.00} & \textbf{1.00} & 0.00 \\ 
  4 & freetext & \textbf{1.00} & \textbf{1.00} & 0.00 & 0.00 & \textbf{1.00} & \textbf{1.00} & \textbf{1.00} & \textbf{1.00} & 0.00 \\ 
  5 & freetext & 0.00 & \textbf{0.55} & 0.18 & \textbf{0.55} & \textbf{0.55} & \textbf{0.55} & \textbf{0.55} & \textbf{0.55} & 0.00 \\ 
  \hline
  6 & dates & 0.00 & \textbf{1.00} & \textbf{1.00} & \textbf{1.00} & 0.00 & \textbf{1.00} & \textbf{1.00} & \textbf{1.00} & 0.00 \\ 
  7 & dates & 0.60 & 0.60 & \textbf{1.00} & 0.28 & 0.60 & 0.60 & 0.60 & 0.60 & 0.00 \\ 
  8 & dates & 0.00 & 0.00 & \textbf{0.91} & 0.00 & 0.00 & 0.00 & 0.00 & 0.00 & 0.00 \\ 
  9 & dates & 0.00 & 0.00 & \textbf{1.00} & 0.00 & 0.27 & 0.00 & 0.00 & 0.00 & 0.00 \\ 
  10 & dates & 0.00 & 0.00 & 0.00 & 0.00 & 0.00 & 0.00 & 0.00 & 0.00 & 0.00 \\ 
  11 & dates & 0.00 & \textbf{1.00} &\textbf{ 1.00} & \textbf{1.00} & 0.00 & 0.00 & \textbf{1.00} & \textbf{1.00} & 0.00 \\ 
  12 & dates & 0.00 & 0.00 & 0.00 & 0.00 & 0.00 & 0.00 & 0.00 & 0.00 & 0.00 \\ 
  13 & dates & 0.00 & 0.00 & 0.00 & 0.00 & 0.00 & 0.00 & 0.00 & 0.00 & 0.00 \\ 
  \hline
  14 & emails & 0.00 & 0.04 & 0.04 & \textbf{1.00} & 0.04 & 0.04 & 0.04 & 0.04 & 0.00 \\ 
  15 & emails & 0.00 & 0.00 & 0.00 & \textbf{1.00} & 0.00 & 0.00 & 0.00 & 0.00 & 0.00 \\ 
  \hline
  16 & names & 0.00 & 0.00 & 0.00 & 0.00 & \textbf{0.91} & 0.00 & 0.00 & 0.00 & 0.00 \\ 
  17 & names & 0.00 & 0.00 & 0.00 & 0.00 & \textbf{0.91} & 0.00 & 0.00 & 0.00 & 0.00 \\ 
  18 & names & 0.00 & 0.00 & 0.00 & 0.00 & \textbf{1.00} & 0.00 & 0.00 & 0.00 & 0.00 \\ 
  19 & names & 0.45 & 0.73 & 0.45 & 0.73 & \textbf{1.00} & 0.73 & 0.73 & 0.73 & 0.00 \\ 
  \hline
  20 & phones & 0.00 & 0.00 & 0.00 & 0.00 & 0.00 & \textbf{1.00} & 0.00 & 0.00 & 0.00 \\ 
  21 & phones & 0.00 & 0.00 & 0.00 & 0.00 & 0.00 & 0.00 & 0.00 & 0.00 & 0.00 \\ 
  22 & phones & 0.00 & 0.00 & 0.00 & 0.00 & 0.00 & \textbf{1.00} & 0.00 & 0.00 & 0.00 \\ 
  23 & phones & 0.00 & 0.27 & 0.00 & 0.27 & 0.27 & \textbf{1.00} & 0.27 & 0.27 & 0.00 \\ 
  24 & phones & 0.00 & \textbf{1.00} & \textbf{1.00} & \textbf{1.00} & 0.00 & \textbf{1.00} & 0.00 & \textbf{1.00} & 0.00 \\
  \hline
  25 & times & 0.36 & 0.91 & 0.91 & 0.91 & 0.91 & 0.91 & \textbf{1.00} & 0.91 & 0.00 \\ 
  26 & times & 0.00 & 0.00 & 0.00 & 0.00 & 0.00 & 0.00 & \textbf{1.00} & 0.00 & 0.00 \\ 
  27 & times & 0.00 & 0.00 & 0.00 & 0.00 & 0.00 & 0.00 & 0.00 & 0.00 & 0.00 \\ 
  28 & times & 0.00 & 0.00 & 0.00 & 0.00 & 0.00 & 0.00 & \textbf{1.00} & 0.00 & 0.00 \\ 
  29 & times & 0.00 & 0.00 & 0.00 & 0.00 & 0.00 & 0.00 & 0.00 & 0.00 & 0.00 \\ 
  \hline
  30 & units & 0.64 & 0.18 & 0.18 & 0.73 & 0.18 & 0.18 & 0.18 & \textbf{1.00} & 0.00 \\ 
  31 & units & 0.00 & 0.00 & 0.00 & 0.00 & 0.00 & 0.00 & 0.00 & \textbf{1.00} & 0.00 \\ 
  32 & units & 0.00 & 0.00 & 0.00 & 0.00 & 0.00 & 0.00 & 0.00 & 0.00 & 0.00 \\ 
   \hline
\end{tabular}
}
\caption{Accuracy obtained per dataset 
depending on the set of primitives (DSBK) used to train \emph{MagicHaskeller}. The results are obtained with $d_{max}$=4, $n=1$ and a maximum execution time of $120s$. Maximum accuracy values in bold.
}\label{tab:mainresults}
\end{table*}

\begin{table*}[!ht]
	\centering
	\resizebox{0.82\textwidth}{!}{
    \def\arraystretch{0.9}
		\begin{tabular}{c|l|l|l|l}
			\textbf{id} &  \textbf{input} & \textbf{expected output} & \textbf{FlashFill} & \textbf{DSI} \\ 
			\hline
			\hline
			
			\multirow{4}{*}{3} & \textit{1-452-789-4567} & \textit{14527894567} &  &  \\
			& 1-406-789-1562 & 14067891562 & {\color{verde} 14067891562} & {\color{verde} 14067891562}  \\
			& 1-4565 & 14565 & {\color{verde} 14565} & {\color{verde} 14565}  \\
			& Etiam dapibus. & Etiamdapibus & {\color{red} } & {\color{verde} Etiamdapibus}  \\
			\hline
			\multicolumn{3}{r|}{Accuracy:} & 0.48 & 1 \\ \hline
			
			\multirow{4}{*}{4} & \textit{International Business Machines} & \textit{IBM} &  &  \\
			& Principles Of Programming Languages & POPL & {\color{verde} POPL} & {\color{verde} POPL}  \\
			& International Conference on Data Mining series & ICDM & {\color{verde} ICDM} & {\color{verde} ICDM}  \\
			& Association of Computational Linguistics & ACL & {\color{verde} ACL} & {\color{verde} ACL}  \\
			\hline
			\multicolumn{3}{r|}{Accuracy:} & 1 & 1 \\ \hline
			
			\multirow{4}{*}{8} & \textit{3/29/86} & \textit{29th} &  &  \\
			& 10 12 69 & 10th & {\color{red} 12th} & {\color{verde} 10th}  \\
			& 04/05/99 & 04th & {\color{red} 05th} & {\color{verde} 04th}  \\
			& 27/07/2007 & 27th & {\color{red} 07th} & {\color{verde} 27th}  \\
			\hline
			\multicolumn{3}{r|}{Accuracy:} & 0 & 1 \\ \hline

			\multirow{4}{*}{9} & \textit{2 of September of 2010,  Monday} & \textit{September} &  &  \\
			& 13 November 2008 & November & {\color{red} 2008} & {\color{verde} November}  \\
			& Tuesday, September 16, 1986 & September & {\color{verde} September} & {\color{verde} September}  \\
			& February 4, 2008 & February & {\color{red} 2008} & {\color{verde} February}  \\
			\hline
			\multicolumn{3}{r|}{Accuracy:} & 0.36 & 1 \\ \hline
            
            \multirow{4}{*}{14} & \textit{Nancy.FreeHafer@fourthcoffee.com} & \textit{fourthcoffee.com} &  &  \\
			& iabetrae@yahoo.es & yahoo.es & {\color{verde} yahoo.es} & {\color{verde} yahoo.es}  \\
			& Sb.edhxo.sk8@hotmail.com & hotmail.com & {\color{verde} hotmail.com} & {\color{verde} hotmail.com}  \\
			& dala\_aguera\_m500@hotmail.com & hotmail.com & {\color{verde} hotmail.com} & {\color{verde} hotmail.com}  \\
			\hline
			\multicolumn{3}{r|}{Accuracy:} & 1 & 1 \\ \hline
            
            \multirow{4}{*}{15} & \textit{Sophia \& domain} & \textit{Sophia@domain.com} &  &  \\
			& elizabeth \& gmail & elizabeth@gmail.com & {\color{verde} elizabeth@gmail.com} & {\color{verde} elizabeth@gmail.com}  \\
			& joypao \& hotmail & joypao@hotmail.com & {\color{verde} joypao@hotmail.com} & {\color{verde} joypao@hotmail.com}  \\
			& casper \& canal13 & casper@canal13.com & {\color{verde} casper@canal13.com} & {\color{verde} casper@canal13.com}  \\
			\hline
			\multicolumn{3}{r|}{Accuracy:} & 1 & 1 \\ \hline
            
            \multirow{4}{*}{17} & \textit{Damian Gobbee} & \textit{D.Gobbee} &  &  \\
			& Antonio Hege & A.Hege & {\color{verde} A.Hege} & {\color{verde} A.Hege}  \\
			& Damancio Hivser-Kleiner & D.Hivser-Kleiner & {\color{red} D.Kleiner} & {\color{verde} D.Hivser-Kleiner}  \\
			& Prof. Edward Davis & E.Davis & {\color{red} P.Davis} & {\color{verde} E.Davis}  \\
			\hline
			\multicolumn{3}{r|}{Accuracy:} & 0.63 & 0.91 \\ \hline
            
            \multirow{4}{*}{19} & \textit{Dr. B. Schdur} & \textit{Dr.} &  &  \\
			& Prof. H. Huifen & Prof. &{\color{verde} Prof.} & {\color{verde} Prof.}  \\
			& Louis Johnson,  PhD & PhD & {\color{red} Lou} & {\color{verde} PhD}  \\
			& Robert Mills &   & {\color{red} Rob} & {\color{verde} }  \\
			\hline
			\multicolumn{3}{r|}{Accuracy:} & 0.72 & 1 \\ \hline
            
            \multirow{4}{*}{20} & \textit{235-7654 \& Taiwan} & \textit{(886) 235-7654} &  &  \\
			& 17-455-81-39 \& Spain & (34) 17-455-81-39 & {\color{red} (886) 17-455-81-39} & {\color{verde} (34) 17-455-81-39}  \\
			& 618-4390 \& Panama & (507) 618-4390 & {\color{red} (886) 618-4390} & {\color{verde} (507) 618-4390}  \\
			& 25-613-24-50 \& Chile & (56) 25-613-24-50 & {\color{red} (886) 25-613-24-50} & {\color{verde} (56) 25-613-24-50}  \\
			\hline
			\multicolumn{3}{r|}{Accuracy:} & 0 & 1 \\ \hline
            
            \multirow{4}{*}{23} & \textit{23/11/18 425-785-4210} & \textit{425-785-4210} &  &  \\
			& 425-613-2450 000-000 & 425-613-2450 & {\color{red} 2450 000-000} & {\color{verde} 425-613-2450}  \\
			& [TS]865-000-0000 - 06-23-09 & 865-000-0000 & {\color{red} 06-23-2009} & {\color{verde} 865-000-0000}  \\
			& 17:58-19:29, 425-743-1650 & 425-743-1650 & {\color{verde} 425-743-1650} & {\color{verde} 425-743-1650}  \\
			\hline
			\multicolumn{3}{r|}{Accuracy:} & 0.36 & 1 \\ \hline
            
            \multirow{4}{*}{25} & \textit{08:55 PM CET} & \textit{08:55} &  &  \\
			& 20:15:00 & 20:15:00 & {\color{verde} 20:15:00} & {\color{verde} 20:15:00}  \\
			& 10:05:00 AM & 10:05:00 & {\color{verde} 10:05:00} & {\color{verde} 10:05:00}  \\
			& UTC 21:20 & 21:20 & {\color{red} UTC 21:20} & {\color{verde} 21:20}  \\
			\hline
			\multicolumn{3}{r|}{Accuracy:} & 0.91 & 1 \\ \hline

           \multirow{4}{*}{28} & \textit{01:34:00 \& 5} & \textit{06:34:00} &  &  \\
			& 01:55  \& 5& 06:55 & {\color{verde} 06:55} & {\color{verde} 06:55}  \\
			& 16:15:12 \& 5& 21:15:12 & {\color{red} 06:15:12} & {\color{verde} 21:15:12}  \\
			& 21:20  \& 5& 02:20 & {\color{red} 06:20} & {\color{verde} 02:20}  \\
			\hline
			\multicolumn{3}{r|}{Accuracy:} & 0.10 & 1 \\ \hline
            
            \multirow{4}{*}{30} & \textit{56.77cl} & \textit{cl} &  &  \\
			& 84Kg & Kg & {\color{verde} Kg} & {\color{verde} Kg}  \\
			& 39.88 A & A & {\color{verde} A} & {\color{verde} A}  \\
			& 1nm & nm & {\color{verde} nm} & {\color{verde} nm}  \\
			\hline
			\multicolumn{3}{r|}{Accuracy:} & 1 & 1 \\ \hline
            
            \multirow{4}{*}{31} & \textit{56.77cl} & \textit{Volume} &  &  \\
			& 84Kg & Mass & {\color{red} Volume} & {\color{verde} Mass}  \\
			& 39.88 A & Electricity & {\color{red} Volume} & {\color{verde} Electricity}  \\
			& 1nm & Length & {\color{red} Volume} & {\color{verde} Length}  \\
			\hline
			\multicolumn{3}{r|}{Accuracy:} & 0.10 & 1 \\ \hline
			
			
		\end{tabular}%
	}
	\caption{Example of results obtained (using \emph{MagicHaskeller} as IP core) compared with \emph{FlashFill}. \emph{Output} is the expected output. The first row of each dataset (\emph{id}) is the example given to \emph{FlashFill} and \emph{MagicHaskeller} to learn. {\color{verde}{Green}} and  {\color{red}{Red}} colours mean, respectively, correct and incorrect results. The accuracy is  $correct\_examples/(total\_examples-n)$, where $n=1$.   }
	\label{comparativa-accuracy}
\end{table*}

The results are much better when the right domain is chosen for the problem. Note that putting all domains together ($all$) implies such big a value of $b$ that \emph{MagicHaskeller} could not solve many problems within the maximum time period. In the same way, some specific problems (datasets \#10, \#12, \#13, \#21, \#27, \#29 and \#32) cannot be solved using a $d_{max}$=4 because they need a higher value in order to find the correct solution.It has to be noticed that since all the DSBK contain some functions for string manipulation, many of them can solve problems related to basic string problems (\textit{freetext} domain). Some problems related to specific domains can also be solved by using basic string manipulation functions, therefore, in this case, any DSBK containing these functions is able to solve the problem. For instance, dataset \#6 (\textit{dates} domain)  can be solved by using constants and the \textit{freetext} function $changePunctuationString$, as we have seen in section \ref{sec:domains}. Since these functions are included in other domains not only \textit{dates} has the best accuracy, but also \textit{freetext}, \textit{emails}, \textit{phones}, \textit{times} and \textit{units}.

We have also compared the performance of our DSI approach using \emph{MagicHaskeller} with other data wrangling tools, concretely, 
\emph{FlashFill} \cite{Gulwani:2011}. \emph{Flashfill} works in the same way as our approach, namely, it uses one or more input instances to try to induce a potential solution,  which is then applied to the rest of examples. If no solution is found or the problem at hand is not solvable by \emph{FlashFill}, it returns, respectively, a void function or an error.

Table~\ref{comparativa-accuracy} shows some illustrative outcomes obtained for each dataset and tool\footnote{For the complete description of results see \url{https://github.com/liconoc/DataWrangling-DSI}} as well as the accuracy values for each dataset. 
The first instance (in italic) for each dataset (\emph{input} column) is the one used for inducing the solution in the different tools. Here, we can see some strength and weakness in each tool. For instance, \emph{Flashfill} works fine with emails and some basic string transformations, but it fails when it has to deal with titles or honorific forms in people names, with problems related to phones prefixes or times and when it has to work with dates in different formats. For its part, DSI using \emph{MagicHaskeller} is able to find the correct solution for the problem at hand, even with only one example, although it still has problems with unexpected punctuation marks (for instance in dataset \#17). In summary, the results show that our approach is able to overcome other tools when dealing with data wrangling problems.

\section{Conclusions}\label{sec:conclusion}
In this paper we have adapted a general IP tool to deal with a range of data wrangling problems by using domain-specific background knowledge. 
Given the impact that the size of the library ($b$) and the size of the solution ($d$) have when solving a data manipulation problem, we found a trade-off that produces positive solutions for many datasets. Finding this trade-off and making it work is novel in the context of IP applied for manipulation problems. 
All this is achieved without the need of increasing the number of examples or using feedback from the user, other than the domain. 
Users can also edit and create the domain files in a general-purpose functional programming language, 
making the system more powerful and able to deal with more and more domains. This contrasts with mainstream approaches based on DSLs, where a change of the DSL aiming at covering other domains cannot be done by the user and might require a redesign of the system. 
Furthermore, the experiments show that our DSI-based approach gets better results than DSL-based approaches, such as FlashFill, mainly due to its adaptability to the problem domain by using domain-specific background knowledge (DSBK).
 
This shows that for these repetitive snippets of code that are necessary for data manipulation problems, we can replace some of the tedious programming effort by the selection of libraries or the definitions of proper functions to handle existing or new domains. Functional programming languages, as we have seen, are particularly appropriate for this. In the end, these data wrangling systems over functional programming languages can actually have the effect of truly incorporating automated programming and program synthesis as a toolbox, even if at the level of the generation of small snippets, for these kinds of applications. 

Finally, we provide what might be considered as the first public repository of datasets for testing data wrangling tools. Although there are several approaches and systems in the literature dealing with the issue under consideration here, none of them provide public access, nor a complete description of the datasets used for their evaluation. In this way, the evaluation procedures are not replicable and neither is the data reusable. 
We have collected different problems from the literature and related software, together with a few freshly-generated ones. With all this data, we have generated a variety of datasets for six different domains 
covering different specific problems in each of them. This repository is open and freely available, and it is already being extended with more types of problems and domains.

As future work, we plan to automate the detection of the domain at hand by using machine learning techniques. The idea is to learn a meta-model that is able to automatically select (or suggest) the appropriate DSBK from the features of the problem. Finally, we plan to integrate everything into a more standalone tool or web service that would allow other users or applications to use this approach in a more standard and accessible fashion. In this way, this approach could be transformed into an API to be used with any language, and potentially included in new tools to complement or compete with those discussed in the related work section.

Beyond the application to data wrangling, we see that the effective combination of background knowledge and hypothesis-driven learning is a particularly promising niche where other areas inside or outside AI, or machine learning alone, are having more difficulties, especially in automation and manipulation problems with very few examples.

\clearpage

\section*{References}
\bibliographystyle{elsarticle-num}
\bibliography{biblio.bib}

\end{document}